\documentclass[conference]{IEEEtran}
\IEEEoverridecommandlockouts

\usepackage{cite}
\usepackage{amsmath,amssymb,amsfonts}
\usepackage{algorithmic}
\usepackage{graphicx}
\usepackage{textcomp}
\usepackage{xcolor}
\usepackage{booktabs}
\usepackage{url}
\usepackage[caption=false,font=footnotesize]{subfig}
\usepackage{gensymb}

\def\BibTeX{{\rm B\kern-.05em{\sc i\kern-.025em b}\kern-.08em
    T\kern-.1667em\lower.7ex\hbox{E}\kern-.125emX}}
\begin{document}

\title{Spatio-temporal Causal Learning for Streamflow Forecasting}

\author{
\IEEEauthorblockN{Shu Wan\IEEEauthorrefmark{1}, Reepal Shah\IEEEauthorrefmark{2}, Qi Deng\IEEEauthorrefmark{2}, John Sabo\IEEEauthorrefmark{2}, Huan Liu\IEEEauthorrefmark{1}, K. Sel\c{c}uk Candan\IEEEauthorrefmark{1}}
\IEEEauthorblockA{
    \IEEEauthorrefmark{1}\textit{School of Computing and Augmented Intelligence, Arizona State University} \\
    \{swan, huanliu, candan\}@asu.edu \\
    \IEEEauthorrefmark{2}\textit{The ByWater Institute, Tulane University}\\
    \{rshah3, qdeng1, jsabo1\}@tulane.edu
    }
}

\maketitle

\begin{abstract}
  Streamflow plays an essential role in the sustainable planning and management of national water resources. Traditional hydrologic modeling approaches simulate streamflow by establishing connections across multiple physical processes, such as rainfall and runoff. These data, inherently connected both spatially and temporally, possess intrinsic causal relations that can be leveraged for robust and accurate forecasting. Recently, spatio-temporal graph neural networks (STGNNs) have been adopted, excelling in various domains, such as urban traffic management, weather forecasting, and pandemic control, and they also promise advances in streamflow management. However, learning causal relationships directly from vast observational data is theoretically and computationally challenging. In this study, we employ a river flow graph as prior knowledge to facilitate the learning of the causal structure and then use the learned causal graph to predict streamflow at targeted sites. The proposed model, Causal Streamflow Forecasting (CSF) is tested in a real-world study in the Brazos River basin in Texas. Our results demonstrate that our method outperforms regular spatio-temporal graph neural networks and achieves higher computational efficiency compared to traditional simulation methods. By effectively integrating river flow graphs with STGNNs, this research offers a novel approach to streamflow prediction, showcasing the potential of combining advanced neural network techniques with domain-specific knowledge for enhanced performance in hydrologic modeling.
\end{abstract}


\section{Introduction}\label{intro}
Streamflow forecasting \cite{sharma2021streamflow} is essential for sustainable water resource management. Accurate predictions of streamflow rates are crucial for various applications, including flood forecasting, water supply management, and ecological preservation. Yet, streamflow forecasting presents a significant challenge due to the inherently nonlinear nature of hydrologic processes and the complex spatio-temporal dynamics that govern these systems. These dynamics are influenced by various forcing variables, such as weather conditions, surface topography, soil type, and vegetation.

Spatio-temporal graph neural networks (STGNNs~\cite{yu2017spatio}) have recently received much attention, demonstrating strong performance in modeling spatio-temporal data. STGNNs start by defining a spatio-temporal graph (STG) where each node represents an instance in space and time. Spatial convolution layers are employed to capture spatial correlations among these nodes, while temporal convolution layers are used to model the temporal dependencies across different time steps. This approach has shown great success in multiple challenging real-world applications, such as smart transportation, climate forecasting, and pandemic prediction.

\begin{figure*}[!t]
    \centering
    \includegraphics[width=0.5\linewidth]{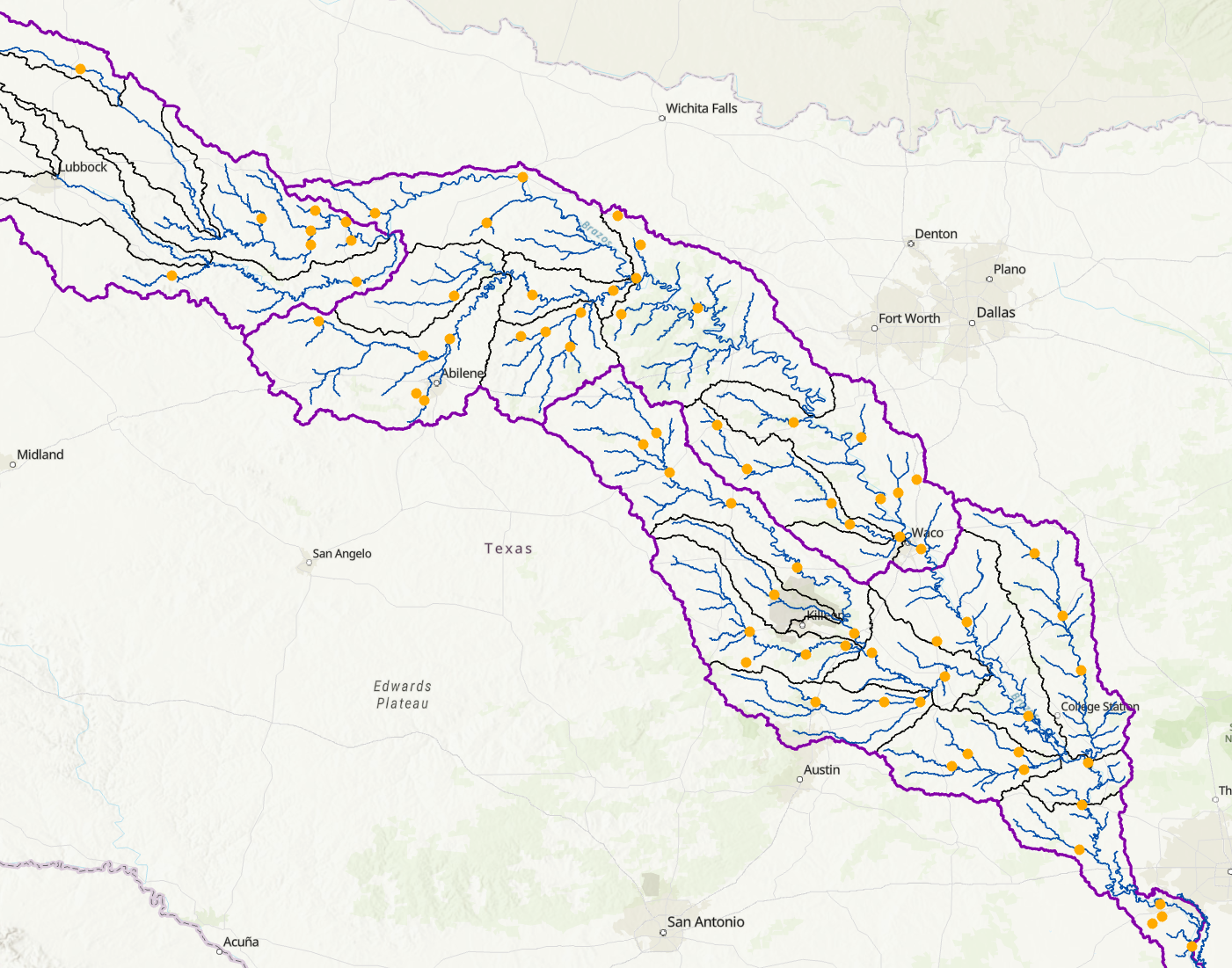}
    \caption{The Brazos River Basin, TX. Each dot represents one monitoring station.}
    \label{fig:brazos}
\end{figure*}

Even though STGNNs have proven effective in modeling spatio-temporal data, this approach faces unique challenges in their application to the domain of hydrology due to their dependence on spatial correlations, making them vulnerable to out-of-distribution shifts. Consider for example the water data analyzed in this study (Figure \ref{fig:brazos}), which comes from the Brazos River basin in Texas: The goal is to forecast future streamflow rates at target stations using the so-called {\em forcing} data from other stations within the basin.
\begin{itemize}
    \item \textbf{Upstream-Downstream Order}: In such scenario,  for accurate streamflow forecasting, it is crucial to consider the directionality of water flow in the river network. For instance, a heavy rainfall event upstream can lead to increased streamflow downstream after a certain time lag. Unfortunately, STGNNs do not incorporate directionality effectively and, consequently, may incorporate information from downstream stations when predicting streamflow at upstream locations, which may be ineffective in this context: for instance, if human activities, such as dam construction, intervene a downstream station, this should not influence the predicted upstream streamflow rates.
    \item \textbf{Drainage Area Subordinations}: The drainage area of a river basin is the region from which all precipitation flows to a common outlet. Stations within the same drainage area are hydrologicly connected, whereas stations from different drainage areas have no direct hydrologic connection. Understanding how different sub-areas contribute to the streamflow at various points in the basin wold help in building more accurate predictive models. Yet, STGNNs might overemphasize spatial proximity in forecasting: For example, they may try to exploit spatial correlations between geographically proximate stations that, in reality, belong to different drainage areas, resulting in inaccurate predictions. 
\end{itemize} 

Machine learning-based spatio-temporal forecasting models typically make predictions by learning a mapping from historical data to future observations without explicitly considering the underlying physical processes. This approach may work for general forecasting tasks but falls short in streamflow forecasting, where understanding and incorporating the physical principles governing water movement within a basin is crucial \cite{nearing2021role}. 
Unlike general spatio-temporal forecasting, streamflow forecasting requires a deeper understanding of the hydrologic connectivity within a basin -- more specifically, understanding the underlying {\em causal relationships},  such as precipitation-runoff relationships and river flow routing mechanics between monitoring stations and forcing variables, is essential for accurate streamflow forecasting.
%
By leveraging knowledge of upstream-downstream hierarchies and drainage area interactions,  forecasting models can more accurately capture the essential causal relationships, leading to more reliable predictions.

To over come these challenges, in this paper we propose \textbf{Causal Streamflow Forecasting} (CSF), a causal-guided spatio-temporal forecasting framework that leverages spatio-temporal causal relationships among nodes while respecting the hydrologic process. Our contributions are outlined as follows:

\begin{itemize}
    \item \textbf{Integration of River Flow Graph}: The CSF model leverages the river flow graph derived from hydrologic principles, using it as prior knowledge to guide spatio-temporal graph convolutional networks (STGCN). This ensures that the model respects the causal relationships between upstream and downstream stations, leading to improved prediction accuracy.
    
    \item \textbf{Hierarchical Network Design}: By incorporating a hierarchical network structure, CSF efficiently handles the complexity of spatio-temporal dependencies in large-scale hydrologic systems. The model clusters stations into groups, reducing computational complexity and enabling efficient parallel training.
    
    \item \textbf{Runoff Embedding Learning}: The first stage of CSF employs a Variational Autoencoder (VAE) to learn latent runoff embeddings that capture the complex relationships between meteorological forcing variables and runoff. This provides an efficient and scalable representation of local hydrologic processes.
    
    \item \textbf{Performance on Real-World Dataset}: We evaluate CSF on a real-world dataset from the Brazos River Basin, demonstrating the model's practical applicability. The results show that CSF consistently outperforms baseline models across short, medium, and long-range forecasting tasks, achieving superior performance.
\end{itemize}

The rest of the paper is organized as follows: in Section~\ref{review}, we briefly introduce the Brazos River Basin and provide a comprehensive review of state-of-the-art methods. In Section~\ref{problem}, we formally define the problem. Section~\ref{approach} presents the proposed approach, including the detailed model architecture, training pipeline, and inference process.
In Section~\ref{experiments}, we describe the evaluation metrics and report the experimental results, followed by further analysis.
Finally, we conclude the paper and highlight future research directions in Section~\ref{conclusion}.

\section{Background and Related Works}\label{review}
In this study, we focus on streamflow forecasting in the Brazos River Basin, the second-largest river basin in Texas, covering a drainage area of 116,000 square kilometers. The basin's namesake river is the 14th longest in the United States, originating from the headwaters at Blackwater Draw in Roosevelt County and flowing to its mouth at the Gulf of Mexico. The Brazos River has the largest average annual flow volume in Texas (Figure~\ref{fig:brazos}).
Streamflow forecasting has evolved significantly over the past decade, including advancements in process-based hydrologic modeling, data-driven machine learning models, physics-informed machine learning techniques to causal-guided machine learning approaches. Each approach offers unique strengths and faces distinct challenges in accurately predicting streamflow.

\begin{figure}[t]
    \centering
    \includegraphics[width=0.9\linewidth]{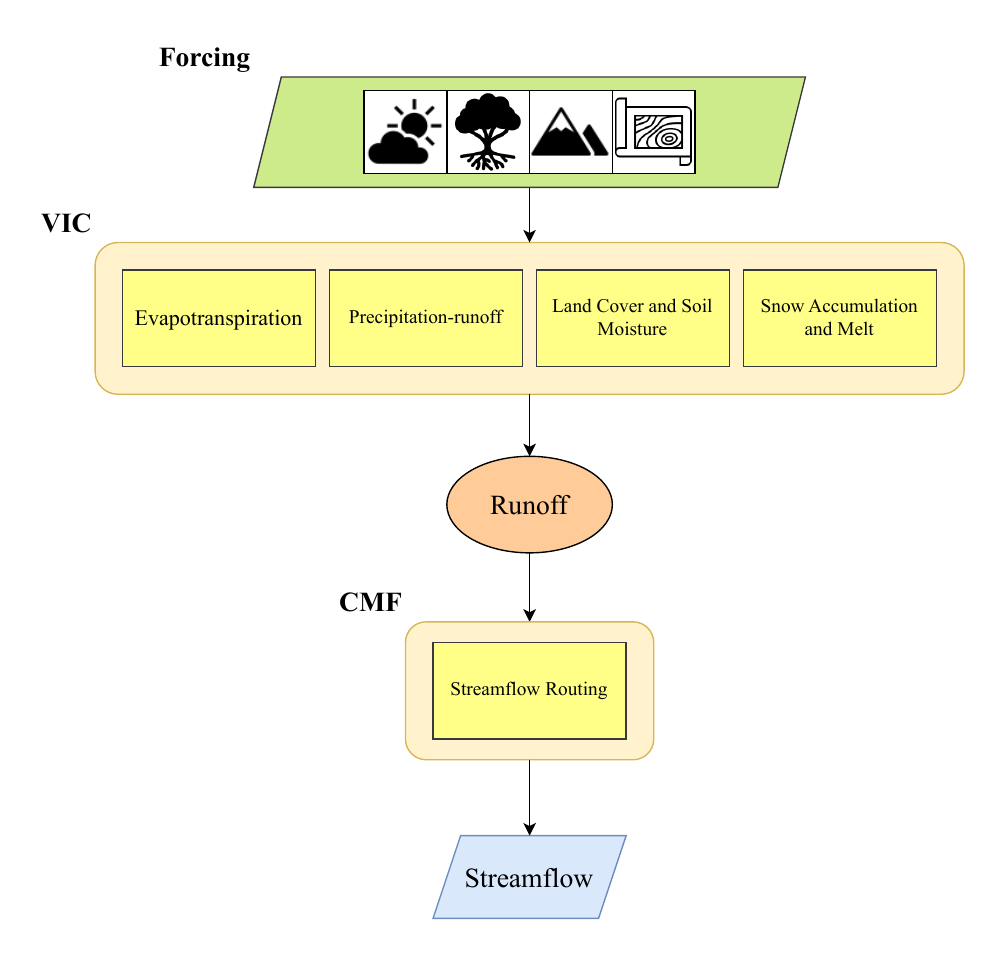}
    \caption{VIC-CaMa-Flood model.}
    \label{fig:vic-cmf}
\end{figure}

\textbf{Process-based hydrologic models} are traditional approaches that simulate various physical processes in hydrology. The Variably Infiltration Capacity (VIC) model \cite{liang1994simple} is a widely-used, large-scale distributed model that accounts for key hydrologic processes, including precipitation-runoff dynamics, evapotranspiration, soil moisture behavior, and river flow mechanics. VIC can be coupled with hydrodynamic models like the Catchment-based Macro-scale Floodplain (CaMa-Flood) model \cite{yamazaki2011physically} to route runoff and estimate streamflow across entire water basins. The combined VIC-CaMa-Flood model (Figure~\ref{fig:vic-cmf}) has demonstrated high fidelity in past studies \cite{shah2023design, wang2021modeling}. However, despite its ability to deliver accurate simulations, this approach requires time-consuming parameters calibration and detailed localized geophysical data, making it challenging forfor similating large number of scenarios, particularly in real-time forecasting scenarios.

\textbf{Data-driven machine learning models}, on the other hand, exploit the power of modern deep learning models, such as Convolutional Neural Networks, Bayesian Neural Network, Graph Neural Network, Convolutional Long Short-Term Memory (Conv-LSTM) networks, have shown promising performance in capturing the complex spatio-temporal relationships in streamflow data \cite{shi2015convolutional, kratzert2018rainfall, kratzert2019toward, zhu2023spatiotemporal, xiang2022fully, kochkov2023neural, Damavandi_Stampoulis_Sabo_Shah_Huang_Wei_Tsai_Srinivasan_Sinha_Boscovic_Low_2020, damavandi2019accurate, liu2023applying, deng2022deep}. These models can learn from historical data to make predictions without the need for explicit physical process representations. For instance, Kratzert et al. \cite{kratzert2018rainfall} demonstrated the capability of LSTM networks to outperform traditional hydrologic models in several basins.

However, one of the significant limitations of these data-driven approaches is their tendency to overlook the causal relationships between monitoring stations. Unlike physical models that inherently incorporate hydrologic connectivity, machine learning models may fail to respect the upstream-downstream order and drainage area subordinations. This oversight can lead to poor forecasting performance, particularly on unseen data where the model's understanding of the physical system's dynamics is tested \cite{chang2023artificial}.

\textbf{Physics-informed machine learning models} attempt to bridge this gap by integrating physical principles with machine learning techniques. Hybrid models, which combine the strengths of traditional hydrologic models and machine learning, are being explored to improve forecasting accuracy while reducing computational demands \cite{shen2018transdisciplinary}. These hybrid approaches aim to leverage the data-driven capabilities of machine learning while embedding essential hydrologic knowledge to respect causal relationships and improve robustness. Previous efforts in rainfall-streamflow forecasting have largely relied on machine learning models that directly map rainfall inputs to streamflow predictions \cite{kratzert2019toward, xiang2022fully}. While these models have demonstrated reasonable accuracy, they often lack the ability to generalize to unseen conditions and provide limited interpretability \cite{nearing2021role, shen2018transdisciplinary}. This shortcoming stems from their failure to explicitly differentiate between the fundamental physical processes involved: the universal laws governing rainfall-runoff generation at each spatial grid and the distinct spatial interactions characterizing runoff routing. An ideal streamflow forecasting model should not only capture the complex relationships between rainfall and streamflow but also disentangle these underlying physical mechanisms to enhance both predictive performance and physical understanding.

\textbf{Causal-guided machine learning models} focus on identifying and leveraging causal relationships in STGs for downstream tasks. For instance, in a river system, the direction of river flow and the connectivity of sub-drainage basins are essential for accurate streamflow forecasting. However, most STGNN methods rely on geo-distance-based adjacency matrices, which can be misleading. Stations close in geo-location may belong to different sub-drainage areas, making such methods prone to inaccuracies. In recent years, causal-guided models have gained increasing attention in spatio-temporal forecasting.
Based on the availability of ground-truth causal STGs, causal-guided models can be classified into two categories. In \cite{deng2023spatio}, the authors define a causal graph for bike flow prediction, analyzing causal relationships between input data, contextual conditions, spatio-temporal states, and predictions. They utilize a representation learning model to learn these representations and use the causal graph to constrain relationships. Similarly, Xia et al. \cite{xia2024deciphering} employ back-door and front-door adjustments to mitigate biases caused by temporal environment and spatial context.
When ground-truth causal STGs are unavailable, causal discovery becomes necessary. For example, in \cite{sheth2022causal}, the authors utilize traditional causal discovery algorithm to first learn a daily causal graph and then using causal community detection module to identify stable feature relations among meteorological forcing variables which are used for predictions. In \cite{sheth2022stcd}, the STCD method extends the TCDF algorithm \cite{nauta2019causal} by incorporating a spatial adjacency matrix based on elevation. Following this work, \cite{sheth2023streams} propose using reinforcement learning to refine causal discovery.
Despite the growing interest, the theoretical framework for causal STGs remains limited due to the complexity of spatio-temporal graphs \cite{christiansen2022toward}. Additionally, the scarcity of ground-truth data in spatio-temporal settings \cite{tec2023space} continues to hinder the development of causal-guided spatio-temporal models.

\begin{figure*}[t]
\centering
\subfloat[\textbf{Station-level model:} The station-level model is a VAE designed to learn an embedding of the forcing variables. Conceptually, it mimics the learning process of traditional process-based models like VIC. The learned runoff embedding does not represent actual runoff; instead, it serves as a latent representation that is passed to the basin-level model in the next stage for further processing.]{\includegraphics[width=0.45\textwidth]{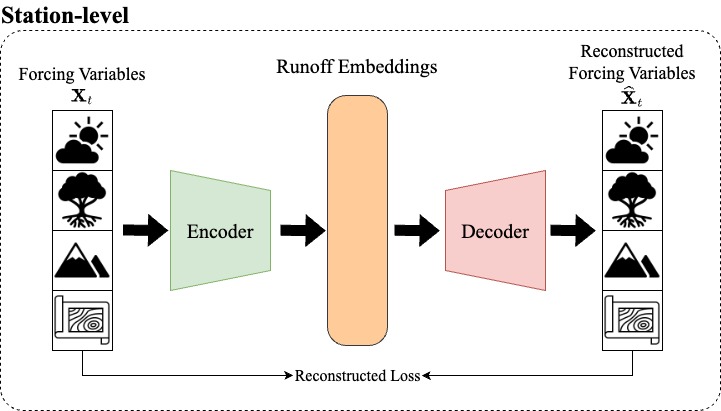}%
\label{fig:station-level}}
\hfil
\subfloat[\textbf{Basin-level model:} The runoff embeddings generated by the station-level model are passed as node embeddings in the basin-level Spatio-Temporal Graph Convolutional Network (STGCN). The river flow graph is constructed using DEM and HUC data, and serves as the spatial causal adjacency matrix to guide causal message passing within the model.]{\includegraphics[width=0.45\textwidth]{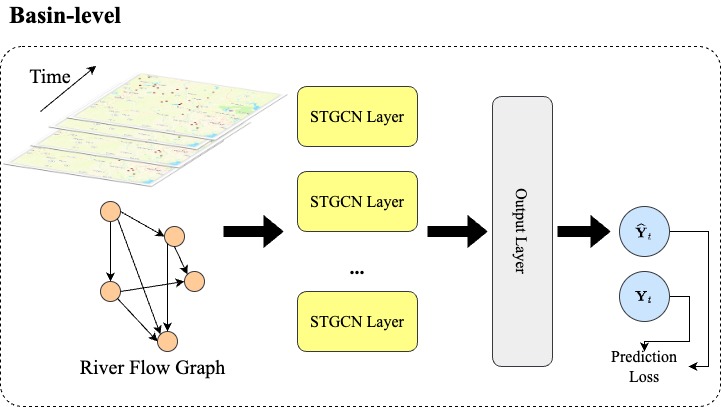}%
\label{fig:basin-level}}
\caption{\textbf{C}ausal \textbf{S}treamflow \textbf{F}orecasting (CSF) model diagram.}
\label{fig:model}
\end{figure*}

Our work differs from previous research in several key ways. First, compared to process-based models that simulate each hydrological process separately, our model learns runoff as a latent embedding that integrates all forcing variables simultaneously. Additionally, unlike previous causal-guided models where causal STGs are either defined by users or discovered through causal discovery algorithms, the river flow graph we use is derived from hydrologic theory and can be considered ground truth. In this work, we utilize the river flow graph as prior knowledge to guide the STGNN, allowing it to aggregate information from causally relevant upstream stations. This approach not only improves interpretability but also enhances scalability in streamflow forecasting.

\section{Problem Formulation}\label{problem}
In this section, we develop the necessary notations and formally define the spatio-temporal graph forecasting problem, followed by the introduction of the causal adjacency matrix used in our model.

\subsection{Spatio-temporal Graph Forecasting}
Spatio-temporal graph forecasting aims to predict future values at multiple spatial locations over time by modeling both spatial and temporal dependencies. Formally, we define a STG as a graph $ \mathcal{G} = (\mathcal{V}, \mathcal{E}) $, where $ \mathcal{V} = \{v_1, v_2, \dots, v_n\} $ represents the set of nodes, and $ \mathcal{E} $ is the set of edges representing relationships between these nodes. The spatial dependencies among the nodes are encoded in the adjacency matrix $ \mathbf{A} \in \mathbb{R}^{n \times n} $, where a non-zero element $ a_{ij} $ indicates a direct connection between nodes $ v_i $ and $ v_j $.

Let $ \mathbf{X}_t = [\mathbf{x}_{1,t}, \mathbf{x}_{2,t}, \dots, \mathbf{x}_{n,t}] \in \mathbb{R}^{n \times d} $ denote the dynamic feature matrix at time step $ t $, where $ \mathbf{x}_{i,t} \in \mathbb{R}^d $ represents the dynamic features of node $ i $ at time $ t $, such as precipitation, temperature, and land use data. Additionally, let $ \mathbf{S} = [\mathbf{s}_1, \mathbf{s}_2, \dots, \mathbf{s}_n] \in \mathbb{R}^{n \times p} $ represent the static feature matrix, where $ \mathbf{s}_i \in \mathbb{R}^p $ corresponds to the static features of node $ i $, such as elevation and soil type.

In this study, we introduce a hierarchical grouping structure. Each node $ i \in \{1, 2, \dots, n\} $ is assigned to a specific group within a hierarchy $ G = \{g_1, g_2, \dots, g_m\} $, where the function $ Group(i) = g_j $ assigns node $ i $ to group $ g_j $. This hierarchical grouping captures sub-drainage areas or regional hydrological divisions that play a critical role in streamflow dynamics.

The goal of spatio-temporal forecasting in this study is to predict future streamflow rates $ \widehat{\mathbf{Y}}_{t+1} $ based on past streamflow rates $\mathbf{Y}_{t-T:t}$, historical dynamic observations $ \mathbf{X}_{t-T:t} $ and static features $ \mathbf{S} $, over a time window $ T $. The spatial dependencies encoded in the adjacency matrix $ \mathbf{A} $ and the hierarchical group structure $ G $ are also considered in the prediction function, as shown in Eq.~\eqref{eq:forecast}:

\begin{equation}\label{eq:forecast}
    \widehat{\mathbf{Y}}_{t+1} = f(\mathbf{Y}_{t-T:t}, \mathbf{X}_{t-T:t}, \mathbf{S}, \mathbf{A}, G),
\end{equation}

\subsection{Causal Adjacency Matrix}
The above problem formulation relies on an adjacency matrix as input. In this work, we construct this adjacency matrix in a causally informed manner. In particular, the causal adjacency matrix $ \mathbf{A} $ is based on the river flow graph generated based on the hydrologic principles. A non-zero entry $ a_{ij}^{\text{causal}} $ in $ \mathbf{A} $ indicates station $ v_i $ is upstream and has direct connection to station $ v_j $. This design ensures that the model respects the directionality and drainage hierarchy within the basin.

\section{Proposed Approach}\label{approach}
The proposed model, Causal Streamflow Forecasting (CSF), involves a two-stage physically-aware hierarchical network designed to integrate both local and global hydrologic processes for accurate streamflow forecasting (Figure~\ref{fig:model}). This approach is divided into two main stages, {\em station-level modeling} and {\em basin-level modeling}, with each stage addressing different aspects of the forecasting problem.

%


\subsection{Two-Stage Physically Aware Hierarchical Modelling}

The two-stage design mimics the traditional process-based VIC-CaMa-Flood model (Figure~\ref{fig:vic-cmf}). The station-level model, acting as a local model, aims to learn a latent runoff embedding. Meanwhile, the basin-level model, serving as a global model, uses a Spatio-temporal Graph Convolutional Network (STGCN) \cite{yu2017spatio} as its backbone and is guided by the river flow graph to learn spatio-temporal dependencies causally.

\subsubsection{Stage 1: Station-level Modeling}
In the first stage, a station-level model is developed to predict runoff using local forcing variables. These variables include weather data (such as precipitation and temperature), soil characteristics, and elevation data. The station-level model captures the dynamic hydrologic processes affecting each monitoring station. We use a Variational Autoencoder (VAE) \cite{kingma2013auto} as the backbone of the station-level model due to its capability of learning complex distributions and capturing the variability in the input data. The latent embedding learned by the VAE is termed the ``runoff embedding" to mimic the runoff process in hydrology. This embedding effectively represents the underlying factors influencing runoff at each station.

A reconstruction loss is employed to guide the training of the VAE, along with the Evidence Lower Bound (ELBO) and regularization terms. The reconstruction loss ensures that the VAE can accurately reconstruct the input forcing variables from the latent space. The ELBO combines the reconstruction loss and a regularization term to ensure the learned latent space follows a predefined prior distribution, typically a standard normal distribution:
\begin{equation}
    \mathcal{L}_{\text{station}} = \mathbb{E}_{q_\phi(\mathbf{z}|\mathbf{x})}[\log p_\theta(\mathbf{x}|\mathbf{z})] - D_{\text{KL}}(q_\phi(\mathbf{z}|\mathbf{x}) \| p(\mathbf{z})),
\end{equation}
where $ \mathbf{x} $ represents the input forcing variables, $ \mathbf{z} $ represents the latent variables (runoff embeddings), $ q_\phi(\mathbf{z}|\mathbf{x}) $ is the approximate posterior distribution, $ p_\theta(\mathbf{x}|\mathbf{z}) $ is the likelihood of the data given the latent variables, $ p(\mathbf{z}) $ is the prior distribution of the latent variables, and $ D_{\text{KL}} $ is the Kullback-Leibler divergence, a regularization term ensuring the latent variables follow the prior distribution.

\begin{table*}[ht!]
\centering
\caption{Model Performance for Short, Medium, and Long Range Forecasting Tasks}
\label{tab:experiment}
\begin{tabular}{@{}lccccccccccccccc@{}}
\toprule
\textbf{Model} & \multicolumn{4}{c}{\textbf{Short Range}} & \phantom{} & \multicolumn{4}{c}{\textbf{Medium Range}} & \phantom{} & \multicolumn{4}{c}{\textbf{Long Range}} \\
\cmidrule{2-5} \cmidrule{7-10} \cmidrule{12-15}
 & \textbf{NSE $\uparrow$} & \textbf{KGE $\uparrow$} & \textbf{VE $\uparrow$} & \textbf{$\rho$ $\uparrow$} & & \textbf{NSE $\uparrow$} & \textbf{KGE $\uparrow$} & \textbf{VE $\uparrow$} & \textbf{$\rho$ $\uparrow$} & & \textbf{NSE $\uparrow$} & \textbf{KGE $\uparrow$} & \textbf{VE $\uparrow$} & \textbf{$\rho$ $\uparrow$} \\
\midrule
MLP      & 0.37 & 0.33 & 0.34 & 0.53 & & 0.25 & 0.26 & 0.29 & 0.31 & & 0.17 & 0.19 & 0.20 & 0.13 \\
CNN      & 0.66 & 0.62 & 0.69 & 0.77 & & 0.55 & 0.57 & 0.53 & 0.61 & & 0.43 & 0.41 & 0.43 & 0.51 \\
Conv-LSTM & 0.68 & 0.61 & 0.71 & 0.79 & & 0.64 & 0.64 & 0.60 & 0.69 & & 0.57 & 0.53 & 0.61 & 0.64 \\
TCDF     & 0.72 & 0.67 & 0.74 & 0.82 & & 0.64 & 0.64 & 0.67 & 0.71 & & 0.60 & 0.59 & 0.65 & 0.64 \\
MTGNN    & 0.74 & 0.71 & 0.73 & 0.63 & & 0.63 & 0.63 & 0.67 & 0.61 & & 0.62 & 0.61 & 0.59 & 0.61 \\
STGCN    & 0.78 & 0.81 & 0.76 & 0.85 & & 0.71 & 0.68 & 0.68 & 0.71 & & 0.67 & 0.57 & 0.67 & 0.67 \\
\midrule
CSF (Our Model)      & \textbf{0.86} & \textbf{0.87} & \textbf{0.88} & \textbf{0.91} & & \textbf{0.82} & \textbf{0.85} & \textbf{0.83} & \textbf{0.88} & & \textbf{0.72} & \textbf{0.67} & \textbf{0.71} & \textbf{0.74} \\
\bottomrule
\end{tabular}
\end{table*}

\subsubsection{Stage 2: Basin-level Modeling}

In the second stage, a basin-level model is utilized to account to capture the broader spatio-temporal dynamics across the entire river basin. 

The model uses a STGCN as the backbone, leveraging the river flow graph as the causal graph. The STGCN is designed to learn the spatio-temporal dependencies between different stations by processing the graph structure that represents the hydrologic connectivity within the basin.


The STGCN operates by first applying spatial graph convolutions followed by temporal convolutions. The spatial convolution at time step $ t $ for node $ i $ can be expressed as:
\[
\mathbf{h}_{i,t}^{(l+1)} = \sigma \left( \sum_{j \in \mathcal{N}(i)} a_{ij} \mathbf{h}_{j,t}^{(l)} \mathbf{W}_s^{(l)} \right),
\]
where $ \mathbf{h}_{i,t}^{(l)} $ is the hidden representation of node $ i $ at layer $ l $, $ \mathbf{W}_s^{(l)} $ is the spatial weight matrix for layer $ l $, $ \mathcal{N}(i) $ is the set of neighboring nodes of $ i $ (determined by $ \mathbf{A} $), and $ \sigma $ is an activation function (e.g., ReLU). This operation aggregates information from causally relevant neighbors for each node.
After the spatial aggregation, temporal dependencies are captured using temporal convolutional layers, which process the hidden representations over time:
\[
\mathbf{h}_{i,t+1}^{(l+1)} = \text{TemporalConv}(\mathbf{h}_{i,t-T:t}^{(l+1)}, \mathbf{W}_t^{(l)}),
\]
where $ \mathbf{W}_t^{(l)} $ is the temporal weight matrix for layer $ l $, and $ \text{TemporalConv} $ represents the temporal convolution operation.
By iteratively applying spatial and temporal convolutions, the STGCN can capture both the spatial relationships defined by the causal adjacency matrix $ \mathbf{A} $ and the temporal dynamics, enabling accurate streamflow forecasting at each node in the network.

Note that, in our design, the runoff embeddings learned in Stage 1 are passed as node embeddings in Stage 2. This integration allows the STGCN to incorporate the localized hydrologic information from each station while learning the broader spatio-temporal relationships across the basin. 

The model's objective is to predict the streamflow rates at each station, accounting for both local and global hydrologic influences. Therefore, the prediction loss is the Mean Squared Error (MSE) between the predicted streamflow rates and the actual observed values:
\begin{equation}
    \mathcal{L}_{\text{prediction}} = \frac{1}{N} \sum_{i=1}^{N} (Y_i - \hat{Y}_i)^2,
\end{equation}
where $ Y_i $ represents the actual streamflow rate at station $ i $, $ \hat{Y}_i $ represents the predicted streamflow rate, and $ N $ is the number of stations. Minimizing this loss function ensures that the predicted streamflow rates closely match the observed values, enhancing the model's accuracy.

The total loss function in the proposed CSF model combines the losses from both the station-level and basin-level stages to optimize the overall performance. The total loss \( \mathcal{L}_{\text{total}} \) is formulated as:
\begin{equation}
    \mathcal{L}_{\text{total}} = \lambda \mathcal{L}_{\text{station}} + (1 - \lambda) \mathcal{L}_{\text{prediction}},
\end{equation}
where \( \lambda \in [0,1] \) is a hyperparameter that controls the relative contribution of each stage's losses.

\textbf{Hierarchical Clustering}. As the number of monitoring stations increases, the complexity of spatio-temporal dynamics grows super-exponentially, making data processing highly demanding. To address this, we leverage the hierarchical structure of stations ($G$) to reduce computational load and improve training efficiency. By clustering stations into hierarchical groups, we adopt a batching approach similar to \cite{chiang2019cluster}, where shuffling occurs at the group level in each epoch. This strategy exploits the limited connectivity between distinct drainage areas, enabling efficient parallel training and reducing the overall complexity of model training.

\subsection{Inference Process}

During inference, the proposed model relies on  causally informed attention to focus on influential stations to predict streamflow rates at target stations. By filtering out irrelevant stations and considering only those with a direct hydrologic impact on the target station, this is done by applying the adjacency matrix $\mathbf{A}$ as a mask. This targeted approach ensures that the predictions are both efficient and reflective of the underlying physical processes governing streamflow dynamics.

\section{Experiments}\label{experiments}

\subsection{Datasets}

The dataset used in this study comes from the Brazos River Basin and consists of five components: the river flow graph, daily meteorological forcings, daily streamflow observations,  sub-drainage area subordination, and simulated runoff.

\begin{figure}[htp]
    \centering
    \includegraphics[width=0.9\linewidth]{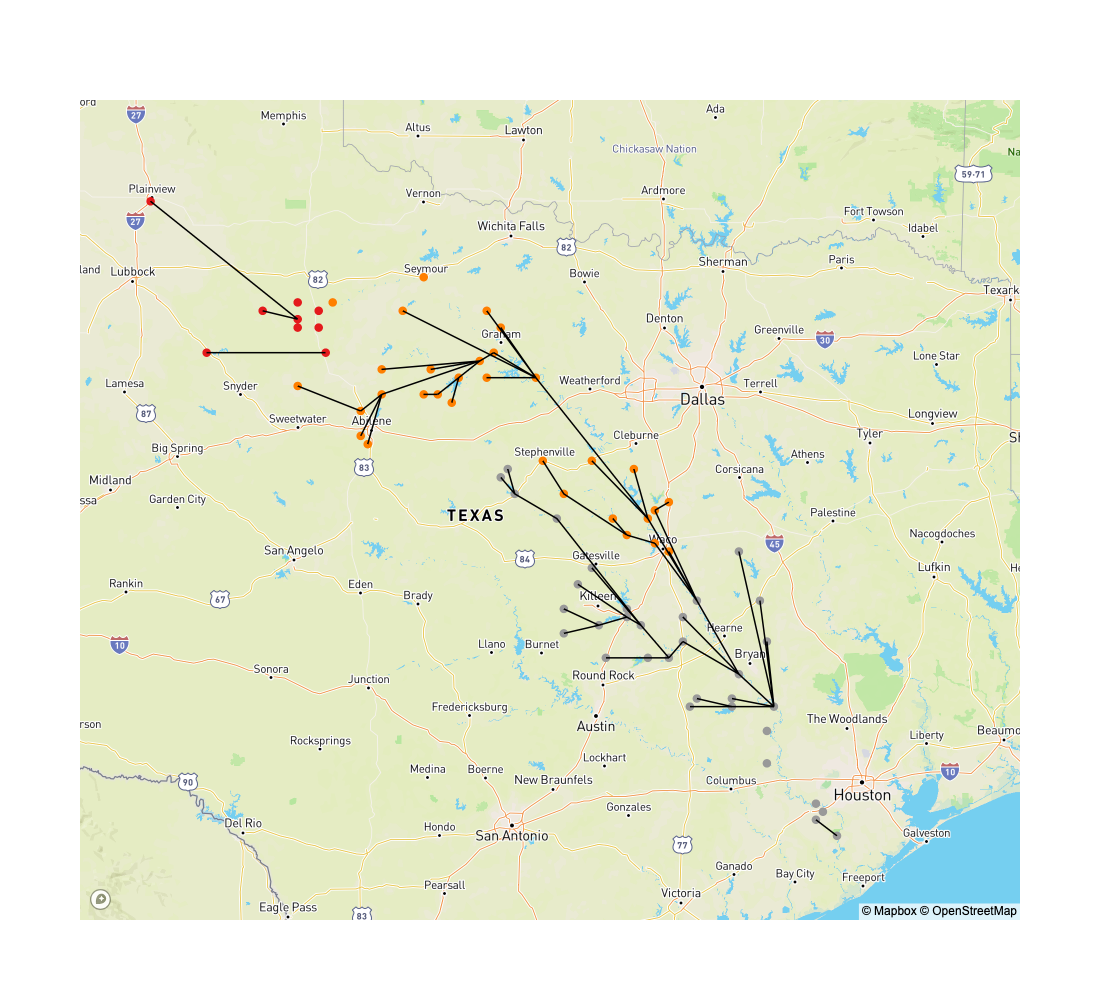}
    \caption{River flow graph. Each color represent a HUC8 subordination.}
    \label{fig:flow_graph}
\end{figure}

The 73 stations in this study are sampled from the whole basin. The river flow graph of the whole basin is generated from Digital Elevation Models (DEM) to capture the upstream-downstream relationships between the stations Figure~\ref{fig:flow_graph}. This graph serves as the basis for constructing the causal adjacency matrix $ \mathbf{A} $, which encodes the directional flow of water between stations.

\begin{figure}[h!]
    \centering
    \includegraphics[width=0.9\linewidth]{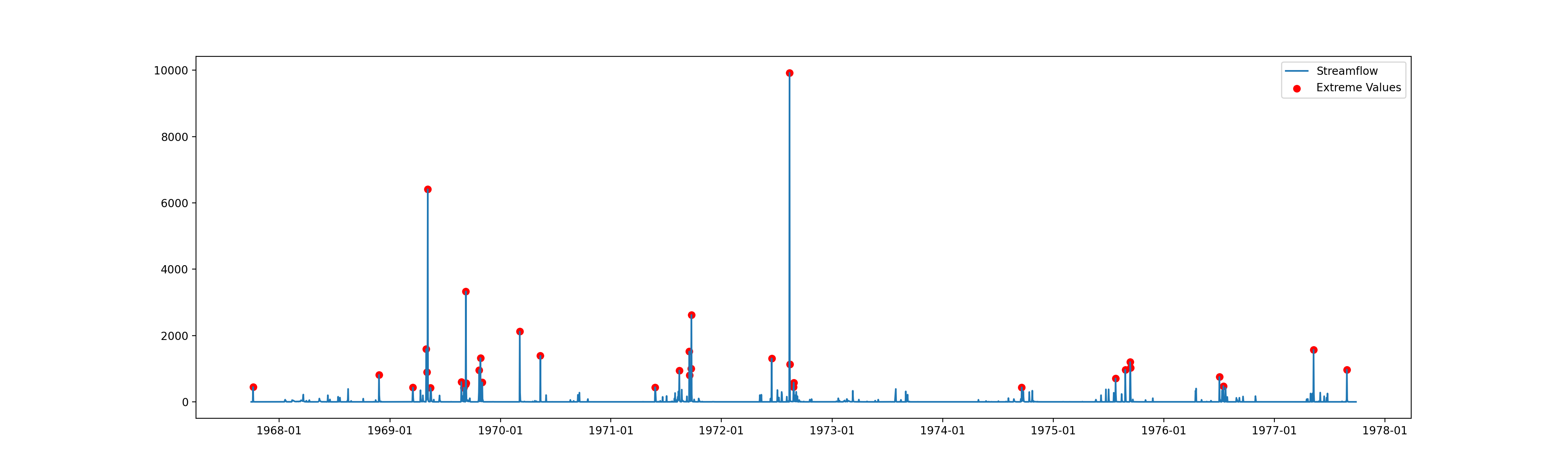}
    \caption{Streamflow data for Grid 3768, with red dots indicating values in the top 0.1 percentile.}
    \label{fig:streamflow}
\end{figure}

The daily meteorological forcings, including precipitation, maximum temperature, minimum temperature, and wind speed, are extracted from Livneh's database \cite{livneh2013long}. This database provides near-surface gridded meteorological and derived hydrologic data for the U.S. continent (CONUS), with a daily temporal resolution spanning from 1915 to 2011, and a spatial resolution of 0.0625 degrees. For this study, we focus on the period from October 1, 1967, to September 27, 1977, comprising a total of 3,650 days. The 73 monitoring stations are located within the bounding box ranging from $29.03125\degree$ to $34.65625\degree$ N and $95.46875\degree$ to $103.84375\degree$ W.

The daily streamflow data is provided by the United States Geological Survey (USGS), with the same temporal resolution as the meteorological forcings. Streamflow data were capped at the 99th percentile to limit the influence of extreme values (Figure~\ref{fig:streamflow}).
The sub-drainage area subordination is determined by the Hydrologic Unit Codes (HUC), provided at HUC8 and HUC4 granularities. Finally, the runoff data is simulated using the VIC model, driven by the same meteorological forcings and calibrated by \cite{shah2023design}.

\subsection{Evaluation}

\subsubsection{Runoff Embedding}
We design CSF as a two-stage physically aware hierarchical model to mimic the VIC-CaMa-Flood model. However, the runoff embeddings $\mathbf{z}$ learned in the station-level model (Figure~\ref{fig:station-level}) are latent and lack direct ground truth for validation. To overcome this issue, we evaluate the learned embeddings with simulated runoff from the VIC model using the \textbf{mutual k-nearest neighbor alignment metric} \cite{klabunde2023similarity}.

Let $ \mathbf{z}_i \in \mathbb{R}^d $ represent the runoff embedding for station $ i $, where $d$ is the embedding size. $ r_i \in \mathbb{R} $ represent the simulated runoff from the VIC model for the same station. For each pair of stations $ (i, j) $, we compute the Euclidean distance between their embeddings and simulated runoffs:

$$d_z(\mathbf{z}_i, \mathbf{z}_j) = \|\mathbf{z}_i - \mathbf{z}_j\|_2, \quad d_r(r_i, r_j) = |r_i - r_j|.$$

For each station $ i $, we define the k-nearest neighbors in terms of Euclidean distance for both the learned runoff embeddings $ \mathbf{z} $ and the simulated runoff $ r $. Let $ \mathcal{S}(\mathbf{z}_i) $ denote the set of indices corresponding to the k-nearest neighbors of $ \mathbf{z}_i $, and $ \mathcal{S}(r_i) $ denote the set of indices for the k-nearest neighbors of $ r_i $.

The alignment metric $ m_{\text{kNN}} $ is then defined as the Jaccard Similarity of the two nearest neighbor sets for each station:

$$m_{\text{kNN}}(\mathbf{z}_i, r_i) = \frac{1}{k} |\mathcal{S}(\mathbf{z}_i) \cap \mathcal{S}(r_i)|$$

The total $k$NN alignment metric is computed as the average of the alignment metrics across all individual stations, capturing how well the learned runoff embedding space aligns with the simulated runoff space. The value ranges between 0 and 1, with higher values indicating better alignment.

\subsubsection{Forecasting performance}
The performance of the proposed model targets several metrics that are common in hydrologic assessment \cite{kim2019hybrid}, including the Nash-Sutcliffe Efficiency (NSE), the Kling-Gupta Efficiency (KGE), Volumetric Efficiency (VE), and the coefficient of determination ($R^2$). These metrics provide comprehensive insights into the accuracy and reliability of the model's predictions.

\textbf{Nash-Sutcliffe Efficiency (NSE)}\cite{nash1970river}: The NSE measures how well the predicted values match the observed data. NSE is defined as
\[
\text{NSE} = 1 - \frac{\sum_{t=1}^{T} (y_t - \hat{y}_t)^2}{\sum_{t=1}^{T} (y_t - \bar{y})^2},
\]
where $ y_t $ is the observed value, $ \hat{y}_t $ is the predicted value, and $ \bar{y} $ is the mean of the observed values. An NSE value of 1 indicates a perfect match, while a value of 0 indicates that the model predictions are as accurate as the mean of the observed data.

\textbf{Kling-Gupta Efficiency (KGE)}\cite{gupta2011typical}: The KGE is designed to provide a more balanced assessment of model performance by considering the correlation, bias, and variability ratio between observed and predicted values. KGE is defined as
\[
\text{KGE} = 1 - \sqrt{(r-1)^2 + (\beta-1)^2 + (\gamma-1)^2}
\]
where $ r $ is the linear correlation coefficient, $ \beta $ is the bias ratio, and $ \gamma $ is the variability ratio. The KGE ranges from -$\infty$ to 1, with higher values indicating better performance.

\textbf{Volumetric Efficiency (VE)}\cite{criss2008nash}: The VE assesses the accuracy of model predictions by calculating the absolute volume difference between observed and predicted values. Unlike NSE, which uses squared differences and can overly penalize large errors, VE uses absolute differences, providing a more balanced view. VE also directly reflects the physical volume of water, making it particularly useful for evaluating total water prediction accuracy. VE is defined as
\begin{equation}
VE = 1 - \frac{\sum |\hat{y}_t - y_t|}{\sum y_t}.
\end{equation}
A VE value of 1 indicates perfect volumetric agreement, while a value closer to 0 indicates significant volumetric mismatch. 

\textbf{Pearson Correlation Coefficient ($\rho$)}: The Pearson correlation coefficient $ \rho $ between the observed values $ y $ and predicted values $ \hat{y} $ is defined as
$$
\rho = \frac{\text{cov}(y, \hat{y})}{\sigma(y) \sigma({\hat{y}})},
$$
where $ \text{cov}(\cdot, \cdot) $ and $ \sigma(\cdot) $ are the covariance and standard deviations operators, respectively. The Pearson correlation coefficient ranges from -1 to 1, with $ \rho = 1 $ indicating a perfect positive linear relationship.

\subsection{Experiment Settings}
To evaluate the performance of CSF, we compare our framework against several baselines, including Multi-Layer Perceptron (MLP), CNN~\cite{lecun1998gradient}, Conv-LSTM~\cite{shi2015convolutional}, TCDF~\cite{nauta2019causal}, MTGNN~\cite{wu2020connecting}, and STGCN~\cite{yu2017spatio}. We design the forecasting tasks with different input and output windows to assess model performance across various temporal ranges. For short-range forecasting, we use a 7-day input window and a 1-day output window. Medium-range forecasting has a 14-day input window with a 3-day output, and long-range forecasting uses a 28-day input window and a 7-day output. 

To ensure consistency across models, we employ a rolling training pipeline that is applied to all baselines and the CSF model. In this rolling pipeline, the model is trained to predict streamflow for the next day, and its prediction is then used as input for forecasting subsequent days.

\begin{table*}[ht!]
\centering
\caption{Ablation Study on CSF key components across different forecasting ranges}
\label{tab:ablation}
\begin{tabular}{@{}lccccccccccccccc@{}}
\toprule
\textbf{Model} & \multicolumn{4}{c}{\textbf{Short Range}} & \phantom{} & \multicolumn{4}{c}{\textbf{Medium Range}} & \phantom{} & \multicolumn{4}{c}{\textbf{Long Range}} \\
\cmidrule{2-5} \cmidrule{7-10} \cmidrule{12-15}
 & \textbf{NSE $\uparrow$} & \textbf{KGE $\uparrow$} & \textbf{VE $\uparrow$} & \textbf{$\rho$ $\uparrow$} & & \textbf{NSE $\uparrow$} & \textbf{KGE $\uparrow$} & \textbf{VE $\uparrow$} & \textbf{$\rho$ $\uparrow$} & & \textbf{NSE $\uparrow$} & \textbf{KGE $\uparrow$} & \textbf{VE $\uparrow$} & \textbf{$\rho$ $\uparrow$} \\
\midrule
Vanilla  & 0.78 & 0.81 & 0.76 & 0.85 & & 0.71 & 0.68 & 0.70 & 0.82 & & 0.59 & 0.55 & 0.57 & 0.67 \\
+ HN       & 0.82 & 0.81 & 0.78 & 0.86 & & 0.73 & 0.71 & 0.73 & 0.83 & & 0.62 & 0.59 & 0.61 & 0.71 \\
+ RG       & 0.84 & 0.86 & 0.85 & 0.90 & & 0.79 & 0.76 & 0.79 & 0.87 & & 0.66 & 0.65 & 0.66 & 0.72 \\
+ HN + RG  & \textbf{0.86} & \textbf{0.87} & \textbf{0.88} & \textbf{0.91} & & \textbf{0.82} & \textbf{0.85} & \textbf{0.83} & \textbf{0.88} & & \textbf{0.72} & \textbf{0.67} & \textbf{0.71} & \textbf{0.74} \\
\bottomrule
\end{tabular}
\end{table*}

\subsection{Results}
\subsubsection{Forecasting Accuracy}

Table \ref{tab:experiment} demonstrates that CSF consistently outperforms all baseline models across short, medium, and long-range forecasting tasks. It achieves the highest scores in all metrics, including NSE, KGE, VE, and $\rho$, indicating superior performance in predicting runoff across varying time horizons. 

The MLP serves as the simplest baseline and performs the worst across all ranges, as it does not consider spatial or temporal dependencies in the data. CNN improves upon MLP by incorporating spatial dependencies, resulting in better performance, particularly in the short range. Conv-LSTM further enhances performance by integrating both spatial and temporal information, showing a noticeable performance jump over CNN. TCDF, MTGNN, and STGCN, which attempt to model STGs, offer incremental improvements by capturing both spatial and temporal dynamics. However, these models do not take into account the specific structure of the river flow graph, which limits their effectiveness compared to CSF. 

CSF’s superior performance can be attributed to its ability to learn the complex dependencies in the hydrological system by modeling both the spatio-temporal dynamics and the actual river flow graph, leading to significant gains across all metrics and forecasting ranges.

\subsubsection{Runoff Embedding Similarities}

\begin{figure}[t]
    \centering
    \includegraphics[width=0.9\linewidth]{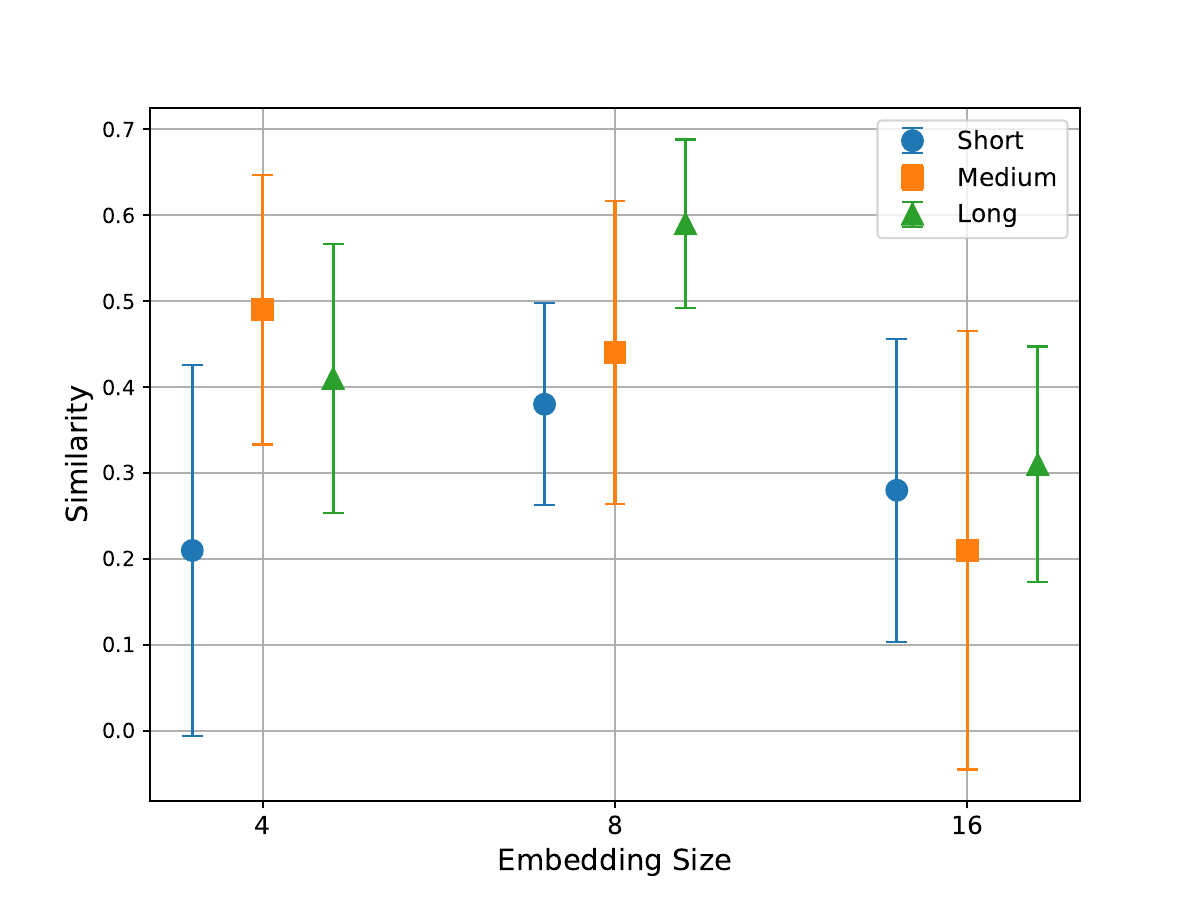}
    \caption{The runoff embedding size versus the $k$NN similarity in different forecasting tasks.}
    \label{fig:emb}
\end{figure}

We conduct this experiment to evaluate the impact of embedding size on the learned runoff embeddings, as it is the key decision in the VAE-based model. Figure \ref{fig:emb} compares embedding sizes of 4, 8, and 16 across short, medium, and long prediction ranges. Embedding size 8 consistently performs well, showing balanced alignment metrics across all ranges. In contrast, embedding size 4 underperforms in capturing longer-term dependencies, and embedding size 16 shows signs of overfitting, particularly in the short range.

From a hydrological perspective, embedding size 8 provides the best trade-off between complexity and generalization. It effectively captures the variability in runoff processes while avoiding overfitting, making it the optimal choice for both short and long-term predictions.

\begin{figure}[htp]
    \centering
    \includegraphics[width=0.9\linewidth]{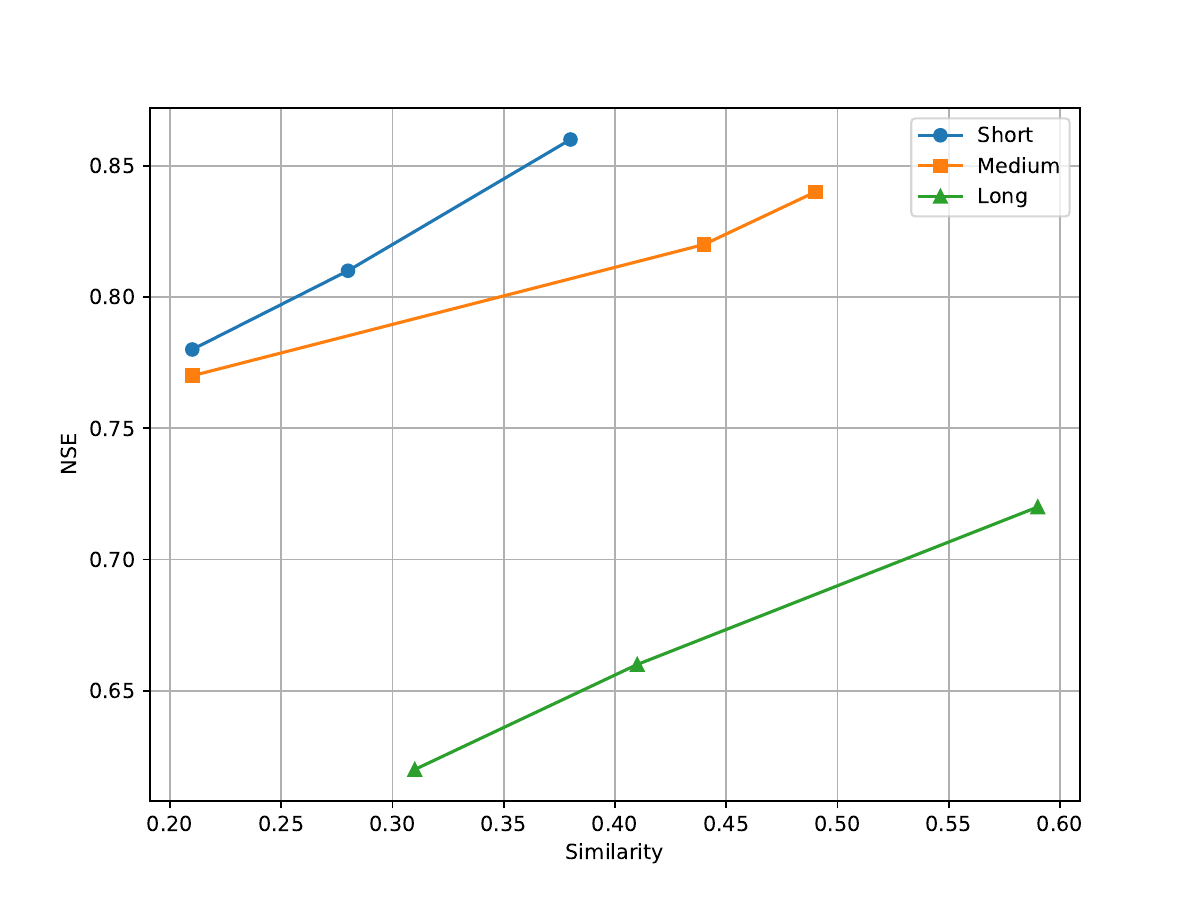}
    \caption{The $k$NN similarity versus NSE in different forecasting tasks.}
    \label{fig:similarity}
\end{figure}

We further analyze the relationship between the similarity of learned embeddings and model performance, as shown in Figure \ref{fig:similarity}. The plot reveals that higher similarity generally correlates with better model performance (NSE), indicating that embeddings with more alignment to the underlying runoff processes improve predictive accuracy. Models trained on longer input windows tend to achieve higher similarity, as they capture more comprehensive hydrological dynamics. However, despite the higher similarity in the long-range models, the NSE is lower compared to shorter ranges due to the increased difficulty of forecasting further into the future, where the forecast window is longer and introduces greater uncertainty in the predictions.

\subsubsection{Ablation Study}
To better understand the contributions of key components in CSF, we conducted an ablation study by removing the hierarchical network (HN) and river graph (RG) modules individually. As seen in Table~\ref{tab:ablation}, the full CSF model, which includes both HN and RG, consistently achieves the best performance across all ranges and all metrics. Removing the HN module results in a noticeable performance drop, especially in long-range forecasting. Similarly, excluding the RG module also reduces the model's performance across all metrics. These results highlight the importance of both the hierarchical structure and the river graph in capturing the spatio-temporal dependencies and improving forecasting accuracy. The combination of these two modules significantly enhances the model's overall effectiveness.

\section{Conclusion, Limitation and Future Work}\label{conclusion}
The CSF model consistently demonstrated superior performance across all forecasting horizons by effectively capturing spatio-temporal dependencies through its integration of the river flow graph and hierarchical network structure. The experiments showed that CSF outperforms traditional and machine learning models, achieving the highest accuracy in key metrics such as NSE, KGE, VE, and $\rho$. The ablation study confirmed the essential role of the hierarchical and river graph components. Embedding size analysis revealed that a balanced representation is key to optimizing model performance. Overall, CSF’s ability to incorporate hydrologic structure and model complex dependencies makes it highly effective for streamflow forecasting, with potential for broader applications in other hydrological settings.

The CSF model has several limitations. First, in its initial stage, the model assumes that stations are independent and identically distributed, hence overlooking spatiotemporal dynamics among stations. Additionally, the forcing-runoff relationship is learned in a hybrid manner using a self-supervised reconstruction loss combined with a supervised forecasting loss. This approach often results in limited generalizability and requires a large volume of training data, where traditional process-based models excel. Moreover, the CSF model is not specifically designed to capture extreme values, which are critical for water management applications. Finally, the model only incorporates four forcing variables focused on rainfall-runoff processes, without considering factors such as vegetation, snowmelt, and human interventions.

Future development will focus on enhancing the model’s capacity to incorporate physical and causal knowledge of water systems, thereby improving its generalizability to unseen data. This includes testing the model on a variety of river basins and developing algorithms that can learn generalized runoff representations that adapt to regional differences in hydrologic conditions.
Moreover, integrating causal discovery in the first stage to understand relationships among forcing variables could improve the accuracy of runoff embeddings and offer deeper insights into underlying hydrological processes.
Lastly, instead of relying on predefined hydrologic knowledge, future work could explore directly learning the river flow graph with minimal supervision. This approach would allow the model to adapt flexibly to different hydrologic systems and provide a data-driven method for capturing spatial dependencies between monitoring stations.

\section*{Acknowledgments}
This research has been funded by NSF\#2311716 and by the US Army Corps of Engineers Engineering With Nature Initiative through Cooperative Ecosystem Studies Unit Agreement \#W912HZ-21-2-0040

\bibliographystyle{IEEEtran}
\bibliography{references}

\end{document}